\documentclass{article}

\usepackage{microtype}
\usepackage{graphicx}
\usepackage{subfigure}
\usepackage{booktabs} 
\usepackage{amsmath}
\usepackage{amssymb}

\usepackage{hyperref}

\usepackage[accepted]{icml2020}

\icmltitlerunning{Graph neural induction of value iteration}

\begin{document}

\twocolumn[
\icmltitle{Graph neural induction of value iteration}



\icmlsetsymbol{equal}{*}

\begin{icmlauthorlist}
\icmlauthor{Andreea Deac}{mila,udem}
\icmlauthor{Pierre-Luc Bacon}{mila,udem}
\icmlauthor{Jian Tang}{mila,hec}
\end{icmlauthorlist}

\icmlaffiliation{mila}{Mila - Quebec AI Institute}
\icmlaffiliation{udem}{University of Montréal}
\icmlaffiliation{hec}{HEC Montréal}

\icmlcorrespondingauthor{Andreea Deac}{andreeadeac22@gmail.com}

\icmlkeywords{Machine Learning, ICML, Graph neural networks, Neural induction, Value iteration}

\vskip 0.3in
]



\printAffiliationsAndNotice{}  

\begin{abstract}

Many reinforcement learning tasks can benefit from explicit planning based on an internal model of the environment. Previously, such planning components have been incorporated through a neural network that partially aligns with the computational graph of value iteration. Such network have so far been focused on restrictive environments (e.g. grid-worlds), and modelled the planning procedure only indirectly. We relax these constraints, proposing a graph neural network (GNN) that executes the value iteration (VI) algorithm, across arbitrary environment models, with direct supervision on the intermediate steps of VI. The results indicate that GNNs are able to model value iteration accurately, recovering favourable metrics and policies across a variety of out-of-distribution tests. This suggests that GNN executors with strong supervision are a viable component within deep reinforcement learning systems.  




\end{abstract}

\section{Introduction} \label{sec:introduction}
The goal of reinforcement learning (RL) \cite{mnih2015human, levine2016end} is to derive a control strategy (a policy) with good long-term behaviour, such as the maximization of returns. The RL framework is highly abstract, encompassing many possible environments, including ones where rewards are very sparse. Sparse rewards imply that many action steps can be performed before any response is given from the environment, making credit assignment (determining the most meaningful actions in a sequence) very challenging for most RL approaches. This motivates the application of \emph{model-based planning} \cite{schrittwieser2019mastering, pascanu2017learning, rivlin2020generalized}, where an explicit model of the environment is used to perform reasoning steps.

Value Iteration Networks (VIN) \cite{tamar2016value} and Generalized Value Iteration Networks (GVIN) \cite{niu2018generalized} propose architectures for deep reinforcement learning that incorporate an explicit planning component. While the planning module partially aligns with the computational graph of the value iteration planning algorithm \cite{bellman1957markovian}, there is no explicit guidance for the network to simulate it -- the entire system is optimised via only a temporal difference loss \cite{sutton1988learning,mnih2015human}. Recent work demonstrated that, when such information is available, steering neural networks towards appropriate step-level outputs can yield tangible benefits for out-of-distribution generalisation and multi-task learning \cite{xu2019can,velivckovic2019neural,yan2020neural}. Inspired by this research direction, our first proposal is directly supervising a neural network on the intermediate steps of value iteration. \\

\begin{figure}
    \centering
    \includegraphics[scale=.3]{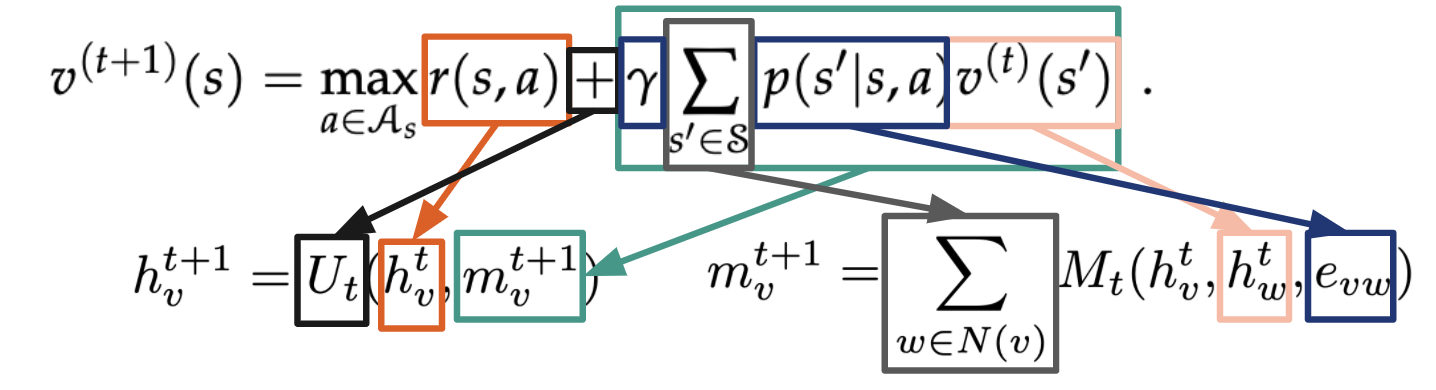}
    \caption{Correspondences between value iteration and graph convolution}
    \label{fig:val_gnn}
\end{figure}

Secondly, VIN is aimed at environments where each state may be represented by a pixel within a grid-world, leveraging the parallels between the computation of value iteration and image convolutions that arise in this setting. This is a clearly restricted setup, as general Markov Decision Process (MDP) states can have a variable number of possible successor states. GVIN generalises the setup to arbitrary graphs, but does not explicitly account for the correspondences between value iteration and the graph convolutional operator---making use of graph kernels \cite{yanardag2015deep} instead. We further make note of the fact that value iteration is a \emph{dynamic programming} algorithm, and it was recently shown that graph neural network (GNN) computations algorithmically align with dynamic programming \cite{xu2019can}. Accordingly, our second proposal is leveraging GNNs that are specially aligned with the computations of the value iteration algorithm \cite{velivckovic2017graph,gilmer2017neural}. \\

\begin{table*}
 \caption{MSE and accuracy for testing different GNNs.}
\centering
 \begin{tabular}{l c c c | c c c}
\toprule 

{\bf Model} & & {\bf MSE} & & & {\bf Accuracy}\\
& $|\mathcal{S}|=20$ &  $|\mathcal{S}|=50$ & $|\mathcal{S}|=100$ & $|\mathcal{S}|=20$ &  $|\mathcal{S}|=50$ & $|\mathcal{S}|=100$ \\
& $|\mathcal{A}|=5$ & $|\mathcal{A}|=10$ & $|\mathcal{A}|=20$ & $|\mathcal{A}|=5$ & $|\mathcal{A}|=10$ & $|\mathcal{A}|=20$ \\

 \midrule   
    MPNN-Sum & 0.457 & 2.175 & 5.154 & 97.75 & 99.3 & 99.32 \\
    MPNN-Mean & 0.455 & 2.199 & 5.199 & 98.125 & 99.3 & 99.32 \\ 
    MPNN-Max & 0.454 & 2.157 & 5.119 & 98. & 99.25 & 99.22 \\
    MPNN-2-Sum & 0.454 & 2.159 & 5.123 & 98.37 & 99.4 & 99.37 \\
    Attn-Sum & 0.757 & 1.725 & 3.765 & 89.75 & 90.55 & 89.69 \\
\bottomrule
 \end{tabular}\label{table:gnn_arch}
\end{table*}

On both fronts, our results are largely positive: we demonstrate that GNNs can be used to simulate value iteration in a manner that generalises to out-of-distribution setups, such as different MDP transition topologies and action/state sizes. GNNs could also enable combining the planning and acting modules -- the setup in this project is the first, model-based, step towards that (assuming perfect knowledge of MDP parameters). However, by considering the MDP model (edge information within the input graph) as information to be derived while also teaching the GNN to compute value iteration, the two modules could be combined towards a model-free algorithm, learning purely from trajectories.

\section{Architecture}
To specify our GNN architecture, we first describe the correspondences between value iteration and graph convolutions---refer to Figure \ref{fig:val_gnn} for a summary.

Firstly, we present the update rule of of value iteration, for an MDP with states $s\in\mathcal{S}$, actions $a\in\mathcal{A}$, transition model $p : \mathcal{S}\times\mathcal{A}\times\mathcal{S}\rightarrow [0,1]$ and reward model $r : \mathcal{S}\times\mathcal{A}\rightarrow\mathbb{R}$: 
\begin{equation}\label{eq:val_iter}
    v^{(t+1)}(s) = \max_{a \in \mathcal{A}_s} r(s,a) + \gamma \sum_{s' \in \mathcal{S}} p(s'|s,a) v^{(t)}(s')
\end{equation}
where $v^{(t)} : \mathcal{S}\rightarrow\mathbb{R}$ is the estimate of $v^*$, the optimal discounted cumulative return, at step $t\in\mathbb{N}$ of the algorithm, and $\gamma\in [0, 1)$ is a discount factor.\\ 

Message passing neural networks (MPNN) \cite{gilmer2017neural} represent the most generic form of graph convolution. They act upon a graph $G=(V,E)$, computing the representations of each node $h_v$ by first computing a vector message from each of its neighbours. They then aggregate all incoming messages into a message vector $m_v$, as follows:
\begin{equation}\label{eq:message}
    m_v^{t+1} = \sum_{w \in \mathcal{N}(v)} M_t(h_v^t, h_w^t, e_{vw})
\end{equation}
where $M$ is the message function, usually a simple MLP. The next representation of node $v$ is then obtained by further transforming the message together with a skip-connection:

\begin{equation}\label{eq:node_update}
    h_v^{t+1} = U_t(h_v^{t}, m_v^{t+1})
\end{equation}
where $U$ is a readout function, usually a simple MLP. The key property satisfied by MPNNs is \emph{permutation invariance}: by aggregating neighbouring messages using a permutation-invariant function (such as summation), we guarantee that the results will be identical under all input graph isomorphisms. Note that other permutation-invariant aggregators can be used in Equation \ref{eq:message}, and we also experiment with maximisation and averaging.\\

For each given MDP, we provide separate graphs $G_{a_i}$ for each action $a_i$, using states $s$ as nodes, and featurise them as follows:
\begin{itemize}
    \item The input node features are defined as follows: $x_s = (v(s), r(s, a_i))$, containing the previous value function estimate and the reward model for taking $a_i$ in $s$;
    \item The input edge features are defined as follows: $e_{s,s'} = (\gamma, p(s'|s,a_i))$, containing the discount factor and the transition model.
\end{itemize}
From this formulation, we can observe the following alignments between value iteration and MPNNs (cf. Figure \ref{fig:val_gnn}):
\begin{itemize}
    \item The MPNN message function $M$ corresponds to taking a product of the edge features with the value-function estimate in the neighbours $s'$;
    \item The aggregation over neighbours corresponds to taking a sum over $s'$ with $p(s'|s,a)>0$; 
    \item The MPNN readout function $U$ correspond to summing the reward model (in the source vertex) with the aggregated messages, in order to add the node's reward to the discounted neighbouring rewards;
    \item The maximisation over actions is aligned with taking the elementwise max over the computed $h_s^{(a_i)}$ within each $G_{a_i}$. The overall vector $h_s = \max_{a_i} h_s^{(a_i)}$ is then used to recompute the next-step value model: $v'(s) = f(h_s)$, where $f$ is an MLP with a single-scalar output.
\end{itemize}

\begin{table*}
 \caption{MSE and accuracy for testing on unseen environments.}
\centering
 \begin{tabular}{l c c c | c c c}
\toprule 

{\bf Model} & & {\bf MSE} & & & {\bf Accuracy}\\
& $|\mathcal{S}|=20$ &  $|\mathcal{S}|=50$ & $|\mathcal{S}|=100$ & $|\mathcal{S}|=20$ &  $|\mathcal{S}|=50$ & $|\mathcal{S}|=100$ \\
& $|\mathcal{A}|=5$ & $|\mathcal{A}|=10$ & $|\mathcal{A}|=20$ & $|\mathcal{A}|=5$ & $|\mathcal{A}|=10$ & $|\mathcal{A}|=20$ \\

 \midrule
    Erd\H{o}s-R\'{e}nyi & 0.457 & 2.175 & 5.154 & 97.75 & 99.3 & 99.32 \\
    Barab\'{a}si-Albert & 0.471 & 2.15 & 5.186 & 98.37 & 99.34 & 99.4 \\
\midrule
    Star & 1.77 & 2.317 & 5.324 & 100. & 100. & 100. \\
    Caveman & 1.452 & 2.302 & 5.285 & 98.12 & 98.84 & 99.05 \\
\midrule
    Caterpillar & 1.039 & 2.674 & 5.474 & 98.37 & 93.34 & 97.05 \\
    Lobster & 0.928 & 2.776 & 5.507 & 97.5 & 92.09 & 95.02 \\
    Tree & 1.01 & 2.672 & 5.494 & 94.25 & 95.59 & 95.19 \\
    Grid & 0.564 & 2.511 & 5.416 & 92.62 & 91.65 & 91.3 \\
    Ladder & 0.605 & 2.536 & 5.487 & 91.25 & 92.15 & 91. \\
    Line & 1.0375 & 2.831 & 5.643 & 88.12 & 88.4 & 89.34 \\

 \midrule
 & $|\mathcal{S}|\approx20$ & & & $|\mathcal{S}|\approx20$  \\
&$|\mathcal{A}|=8$ & & &$|\mathcal{A}|=8$ & & \\
\midrule
Maze \cite{tamar2016value} & 4.95 & & & 69.86 \\
\bottomrule
 \end{tabular}\label{table:graph_types}
\end{table*}

\section{Experiments}
For all experiments performed, we train on MDPs with transition models following\footnote{We define an MDP as \emph{following} a graph distribution when the set of the nonzero entries of its transition model, i.e. $\mathcal{E} = \{(s, s')\ |\ p(s'|s, a_i) > 0\}$, can be described as generated from this distribution.} Erd\H{o}s-R\'{e}nyi graphs \cite{erdos1959random} with 20 states and 5 actions ($|\mathcal{S}|=20$, $|\mathcal{A}|=5$). For each iteration step $t$, we optimise the mean-squared error between the prediction $v'(s)$ and the ground-truth value obtained from value iteration at each state. Training is performed with teacher forcing (always predicting the outputs of one iteration from ground-truth inputs), while at test time we perform rollouts of the GNN, feeding back the predicted values $v'(s)$ as inputs in the next step until convergence.\\

\begin{figure*}
    \hfill
    \includegraphics[width=0.325\linewidth]{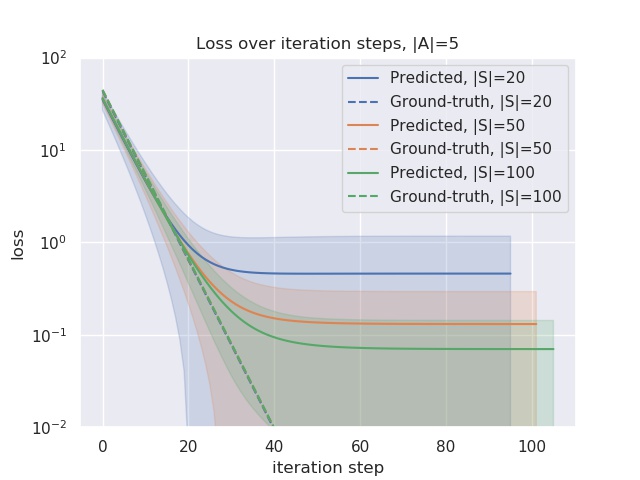}
    \hfill
    \includegraphics[width=0.325\linewidth]{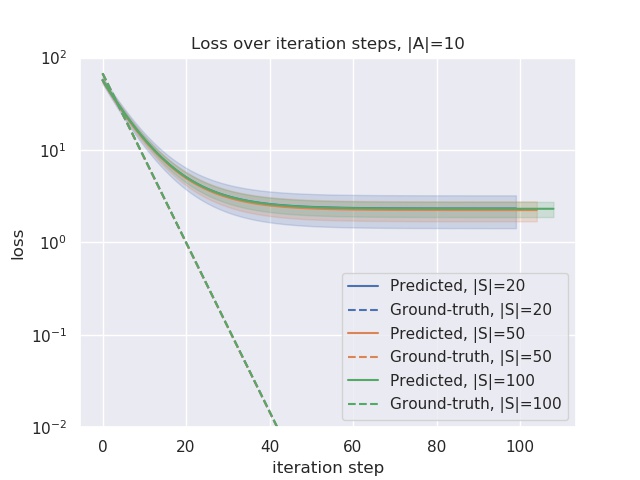}
    \hfill
    \includegraphics[width=0.325\linewidth]{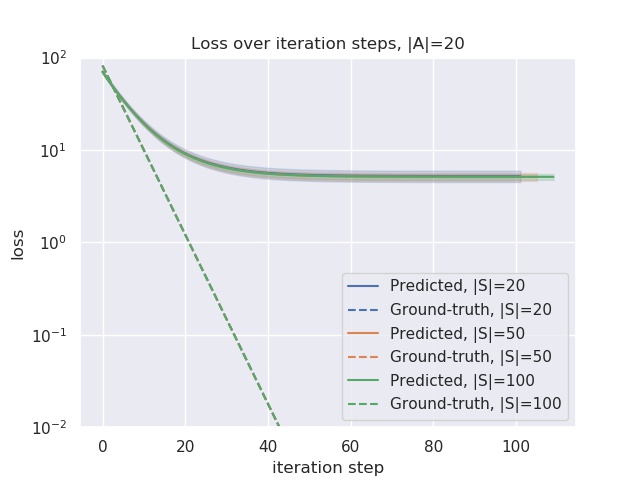}
    \hfill
        \includegraphics[width=0.325\linewidth]{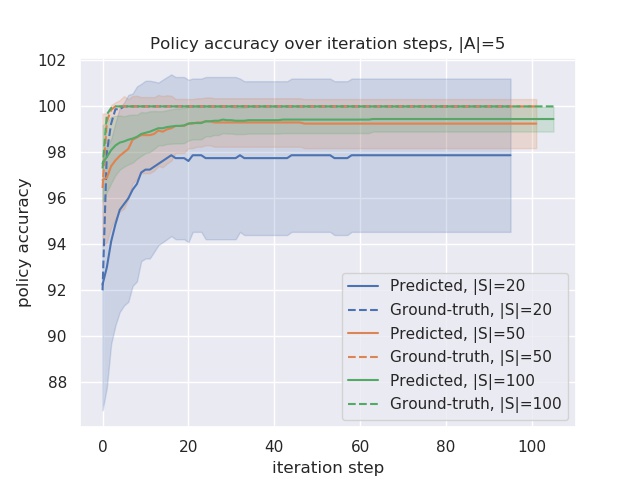}
    \hfill
        \includegraphics[width=0.325\linewidth]{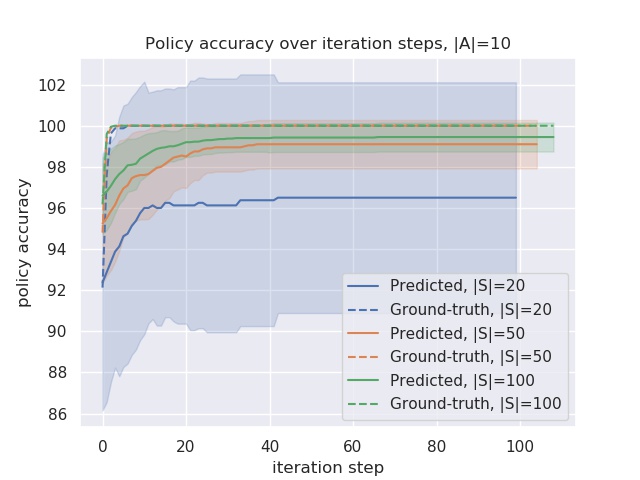}
    \hfill
        \includegraphics[width=0.325\linewidth]{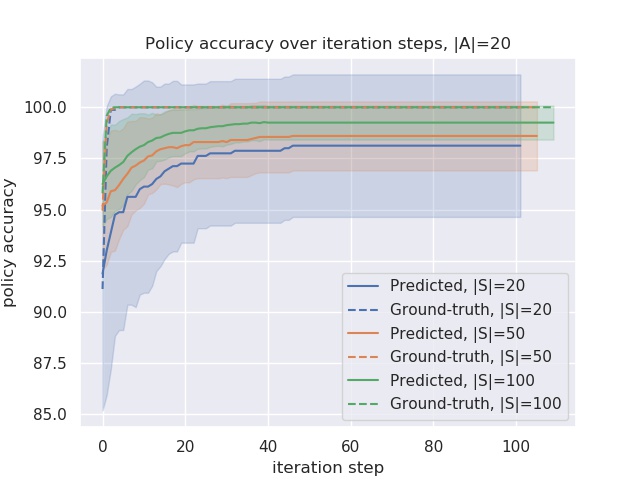}
    \hfill
    \caption{{\bf Fixed $|\mathcal{A}|$ } -- from left to right, the number of actions is 5, 10 and 20 respectively. The x-axis shows the number of iteration steps, while the y-axis represents the MSE in the first row and the accuracy in the second row. \emph{Blue} is $|\mathcal{S}|=20$, \emph{orange} is $|\mathcal{S}|=50$ and \emph{green} is $|\mathcal{S}|=100$. The continuous line corresponds to MPNN-Sum, while the dashed line corresponds to value iteration. }
    \label{fig:fixed_action}
\end{figure*}

Besides checking the MSE on the final values obtained from test rollouts w.r.t. $v^*$, we also compute the \emph{policy accuracy}---checking, for each state, the overlap of the policy obtained by taking the argmax of Equation \ref{eq:val_iter} for the predicted values $v'$ and the optimal value $v^*$. Even if our MPNN model incorrectly estimates $v^*$, the predicted values could still result in identical policies if the relative differences between the entries of $v'$ match the ones of $v^*$. In fact, there are many possible solutions $v$ that respect the underlying Bellman equation in a way that recovers the optimal policy. \\

We aim to check that the model generalises to out-of-distribution samples, learning the VI algorithm and not nuances in the training data: therefore, we conduct out-of-distribution testing on graphs of sizes ($|\mathcal{S}|=50$, $|\mathcal{A}|=10$) and ($|\mathcal{S}|=100$, $|\mathcal{A}|=20$). Starting with the MPNN-Sum model as described in the previous section (using linear layers for $M$, $U$ and $f$), we visualise the generalisation test results for varying state space size $|\mathcal{S}|$ in Figure \ref{fig:fixed_action}. Here, we fix the number of actions and vary number the of states in $\{20, 50, 100\}$, plotting the MSE and policy accuracy against $v^*$ as a function of MPNN iteration count. The results indicate that the model is robust in all cases in terms of recovered policy, and the iterations gracefully converge to a fixed loss. \\

The experiments in Table \ref{table:gnn_arch} explore the particularities in the GNN, in order to confirm that the architecture proposed in Section 2 is indeed appropriate. Besides MPNN-Sum, we also experiment with averaging (MPNN-Mean) and maximisation (MPNN-Max). The results demonstrate, however, that the model is robust to these changes. Moreover, a one-layer message function ($M$) might not have the ability to accurately model the product between $\gamma$ and $p(s'|s,a)$, so we also experiment with a two-layer MLP as $M$ (denoted as MPNN-2-Sum); this yields similar results to MPNN-Sum. Lastly, we note that the message could be used as a multiplicative factor in VI, which aligns it well with attention (Attn-Sum) \cite{velivckovic2017graph} as a way to scale neighbours, rather than using messages. While this operator indeed reduces the overall MSE, it also significantly reduces the recovered policy accuracy. We hypothesise that this is due to the fact that attending has scale-preservation built-in (and hence less chance of over-estimating $v^*$), but is still not powerful enough to model the relative values of $v'$ in the best way.\\

We conclude with a generalisation test on the graph distribution that the MDP follows. After training the MPNN-sum model on Erd\H{o}s-R\'{e}nyi MDPs, we check for generalisation over different environments by testing on MDPs with different underlying graph distributions.\\

The results are enumerated in Table \ref{table:graph_types}, and span the same graph distributions as in \cite{corso2020principal}: from scale-free graphs (Barab\'{a}si-Albert \cite{barabasi1999emergence}), through graphs with densely-connected patterns (Star, Caveman \cite{watts1999networks}) to sparse graphs (Caterpillar, Lobster, Tree, Grid, Ladder, Line). Finally, we evaluate the extent to which our VI executor can zero-shot generalise to a fully deterministic $8\times 8$ maze environment as described in VIN \cite{tamar2016value}.\\

We can observe that the model exhibits a strong level of zero-shot generalisation to most unseen MDP graphs; better so when the graphs are more dense or more closely related to the training distribution. As the graphs get sparser, so do the MSE and policy accuracy degrade---dropping to 70\% in the fully deterministic case. Such results are to be expected, as the MPNN was trained on dense graphs with edges and transitions sampled uniformly at random. If we can anticipate any properties of our test environments at training time (e.g. determinism), we may appropriately modify the training distribution to accommodate this.


\section{Conclusions}
We proposed a method for performing neural induction of value iteration using graph neural networks. By testing on out-of-distribution graph sizes and unseen environments, we show that the algorithm learnt by GNNs is robust w.r.t the number of states and on other types of random graphs. Out-of-distribution performance aligns with the structural similarities between training and testing MDPs in an expected manner, highlighting the potential of training a targeted executor provided we know certain characteristics of our test environments, for example sparsity, degree distribution and determinism. \\

Our work indicates that GNN executors can become a robust component in differentiable planning systems; however, further work remains to be done before they can be broadly applicable. GNNs could also enable the combination of planning and acting: by re-estimating the MDP model parameters from observed trajectory data, simultaneously with learning to execute value iteration, we can derive a model-free general purpose algorithm which combines latent graph inference with differentiable planning.





\bibliography{references}
\bibliographystyle{icml2020}


\end{document}